\documentclass[nonacm,screen,natbib=false]{acmart} 
\AtBeginDocument{%
  }

\setcopyright{acmcopyright}
\copyrightyear{2025}
\acmYear{2025}
\acmDOI{XXXXXXX.XXXXXXX}

\RequirePackage[
  datamodel=acmdatamodel,
  style=acmnumeric,
  ]{biblatex}

\usepackage{comment}
\usepackage{amsmath}
\usepackage{soul}
\usepackage{subcaption}
\usepackage{booktabs} %
\usepackage{multirow}
\usepackage{soul}%
\usepackage{xcolor} %
\usepackage{colortbl}
\usepackage{changepage}
\usepackage{threeparttable}

\newcommand\co{CO\textsubscript{2}}
\newcommand\gcoeq{gCO\textsubscript{2}eq}

\DeclareUnicodeCharacter{0308}{\textcolor{red}{HERE!HERE!08}}
\DeclareUnicodeCharacter{0301}{\textcolor{red}{HERE!HERE!01}}
\begin{document}

\title{Efficiency is Not Enough: A Critical Perspective of Environmentally Sustainable AI}

\author{Dustin Wright}
\email{dw@di.ku.dk}
\orcid{1234-5678-9012}
\affiliation{%
  \institution{University of Copenhagen}
  \city{Copenhagen}
  \country{Denmark}
}

\author{Christian Igel}
\email{igel@di.ku.dk}
\orcid{1234-5678-9012}
\affiliation{%
  \institution{University of Copenhagen}
  \city{Copenhagen}
  \country{Denmark}
}

\author{Gabrielle Samuel}
\email{gabrielle.samuel@kcl.ac.uk}
\orcid{1234-5678-9012}
\affiliation{%
  \institution{Kings College London}
  \city{London}
  \country{United Kingdom}
}

\author{Raghavendra Selvan}
\email{raghav@di.ku.dk}
\orcid{1234-5678-9012}
\affiliation{%
  \institution{University of Copenhagen}
  \city{Copenhagen}
  \country{Denmark}
}

\renewcommand{\shortauthors}{Wright et al.}

\begin{abstract}
  Artificial intelligence (AI) is currently spearheaded by machine learning (ML) methods such as deep learning which have accelerated progress on many tasks thought to be out of reach of AI. These recent ML methods are often compute hungry, energy intensive, and result in significant green house gas emissions, a known driver of anthropogenic climate change. Additionally, the platforms on which ML systems run are associated with environmental impacts that go beyond the energy consumption driven carbon emissions. 
The primary solution lionized by both industry and the ML community to improve the environmental sustainability of ML is to increase the compute and energy efficiency with which ML systems operate. In this perspective, we argue that it is time to look beyond efficiency in order to make ML more environmentally sustainable. We present three high-level {\em discrepancies} between the many variables that influence the efficiency of ML and the environmental sustainability of ML. Firstly, we discuss how compute efficiency does not imply energy efficiency or carbon efficiency. Second, we present the unexpected effects of efficiency on operational emissions throughout the ML model life cycle. And, finally, we explore the broader environmental impacts that are not accounted by efficiency.
These discrepancies show as to \textit{why} efficiency alone is not enough to remedy the adverse environmental impacts of ML. Instead,  we argue for systems thinking as the next step towards holistically improving the environmental sustainability of ML.
\end{abstract}



\maketitle

\section{Introduction}

Artificial intelligence (AI) is rapidly becoming ubiquitous, so much so it has been argued that ``AI [\dots] is becoming an infrastructure that many services of today and tomorrow will depend upon''~\cite{robbins2022our}. Current progress in the field of AI is spearheaded by machine learning (ML) techniques such as deep learning~\cite{lecun2015deep,schmidhuber2015deep}, which has rendered many tasks previously thought to be out of reach of AI more or less solved~\cite{DBLP:conf/nips/BrownMRSKDNSSAA20,jumper2021highly,DBLP:conf/cvpr/RombachBLEO22,silver2016mastering}. Deep learning methods can be characterized as overparameterized function approximators trained to learn from data, where scale, i.e. quantity of data and computational footprint, are often seen to have a positive impact on performance~\cite{thompson2020computational,igel:21KIed}. In line with this, the past decades have seen an exponential rise in the amount of compute used by ML systems~\cite{DBLP:conf/ijcnn/SevillaHHBHV22,desislavov2021compute}, which has led to a subsequent rise in energy consumption and carbon emissions~\cite{desislavov2021compute,DBLP:journals/computer/PattersonGHLLMR22,DBLP:conf/mlsys/WuRGAAMCBHBGGOM22,DBLP:journals/corr/abs-2302-08476}. These carbon emissions come from multiple sources, including operational emissions from direct compute across the ML model life cycle (i.e. the development and deployment of ML systems) and emissions from the supply chain needed to produce ML hardware and cloud data centers (i.e. embodied emissions). Beyond carbon emissions, increased production and use of the hardware infrastructure needed for ML is potentially exacerbating broader environmental impacts, including fresh water consumption for cooling, pollution from e-waste, mining for resources to build ML platforms, and more~\cite{DBLP:journals/corr/abs-2304-03271}. While on the one hand ML systems can be used \textit{for} making progress towards the sustainable development goals (SDGs)~\cite{DBLP:journals/aiethics/Wynsberghe21,DBLP:journals/csur/RolnickDKKLSRMJ23}, on the other hand the above mentioned factors limit the sustainability \textit{of} ML from an environmental perspective.

A major focus of the ML community in pursuit of sustainable ML (more specifically improving the sustainability \textit{of} ML~\cite{DBLP:journals/aiethics/Wynsberghe21}) 
has been to make ML systems and the hardware that runs them more \textit{efficient}~\cite{DBLP:journals/corr/abs-2301-11047,DBLP:conf/mlsys/WuRGAAMCBHBGGOM22,DBLP:journals/computer/PattersonGHLLMR22,DBLP:journals/corr/abs-2210-06640}. Efficiency in this context is understood through the relationship between three factors: {\em compute}, generally measured in terms of floating point operations per-second (FLOPS), the number of parameters used by an ML system, and/or the amount of time needed to perform a particular computation; {\em energy} which is generally measured in terms of kilowatt hours (kWh) required to perform the compute; and {\em carbon}, generally measured in terms of equivalent grams of \co{} (\gcoeq{}) emitted due to the energy consumption. The aim of ML efficiency is to reduce the costs (e.g. energy or carbon) for a given unit of output (e.g.~compute). This means reducing the compute and/or energy consumption of ML systems without sacrificing their utility in the form of e.g. performance on a given set of tasks. These improvements \textit{can} reduce the carbon emissions of ML systems, and should be continued, but they can also fall short. This is evident when considering the overall goal of improving environmental sustainability of ML as improvements in efficiency often have unexpected effects~\cite{DBLP:journals/corr/abs-2002-05651,widdicks2023systems,font2022rebound}. These unexpected effects come in many forms, such as when a reduction in compute (e.g., through neural network sparsification) leads to an increase in carbon emissions (e.g., due to increased energy consumption from inefficient sparse operations) or the use of a more efficient system leads to greater overall use of that system over time. Additionally, efficiency primarily addresses operational emissions while exacerbating the relative impact of embodied emissions, and may be outpaced by the growing infrastructure needed to support ML as a technology~\cite{batra2018artificial,DBLP:conf/mlsys/WuRGAAMCBHBGGOM22,robbins2022our,kaack2022aligning}.

In this paper, we present a critical perspective of environmentally sustainable ML which examines the relationship between the efficiency of ML systems and their overall environmental impact. We focus specifically on efficiency as it relates to the sustainability \textit{of} ML as opposed to ML \textit{for} sustainability which seeks to use ML systems and AI more generally towards reaching the SDGs~\cite{DBLP:journals/aiethics/Wynsberghe21,DBLP:journals/csur/RolnickDKKLSRMJ23}. As such, this perspective synthesizes a large body of research on efficiency and environmental sustainability, both in general~\cite{mensah2019sustainable} and within the sustainability of ML~\cite{kaack2022aligning}. With this we hope to comprehensively demonstrate, at multiple levels of granularity providing both technical and non-technical reasons, \textit{why} efficiency alone is not enough to remedy the adverse environmental impacts of ML.
We express this through three high-level {\em discrepancies} between the effect of efficiency on the environmental sustainability of ML when viewed narrowly and when considering the many variables with which it interacts:

\begin{figure}[t]
  \centering
  \begin{subfigure}[b]{0.46\columnwidth}
    \centering
    \includegraphics[width=0.9\textwidth]{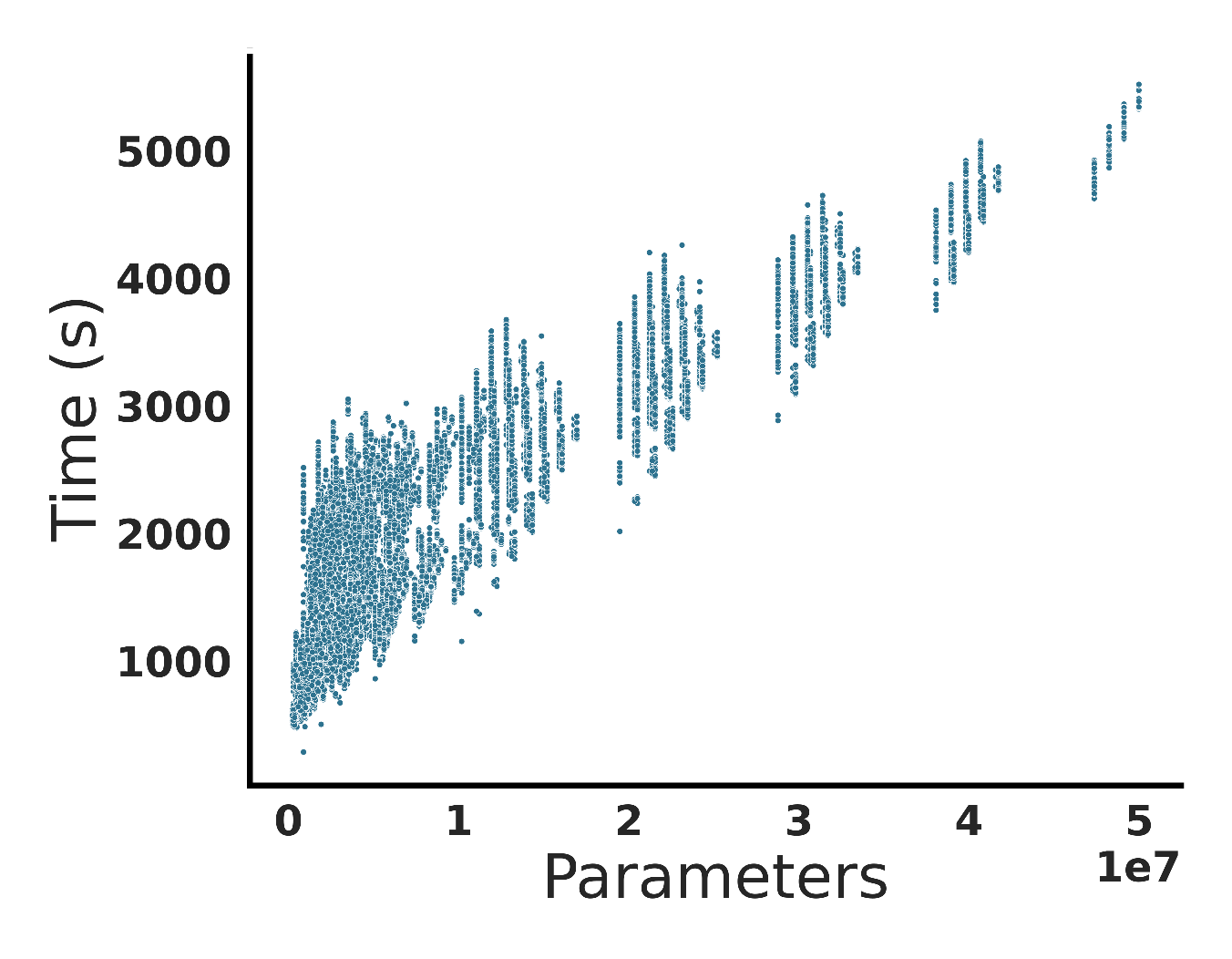}
    \caption{ Training time vs. number of parameters
    }
    \label{fig:time_vs_params}
  \end{subfigure}\hfill
    \begin{subfigure}[b]{0.46\columnwidth}
        \centering
        \includegraphics[width=0.9\textwidth]{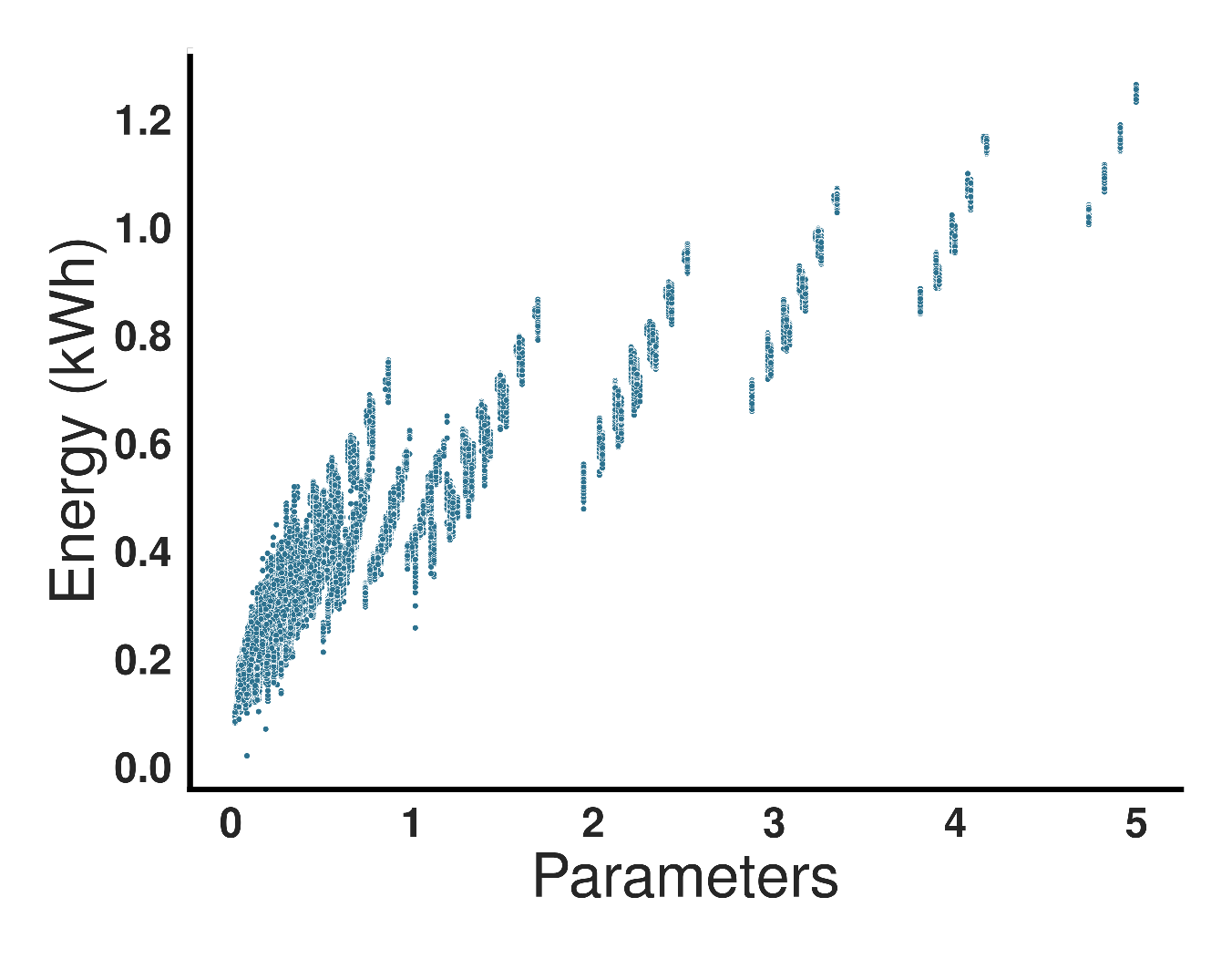}
        \caption{ Energy consumption vs. number of parameters
        }
        \label{fig:energy_vs_params}
    \end{subfigure}
    \\
    \begin{subfigure}[b]{0.46\columnwidth}
        \centering
        \includegraphics[width=0.9\textwidth]{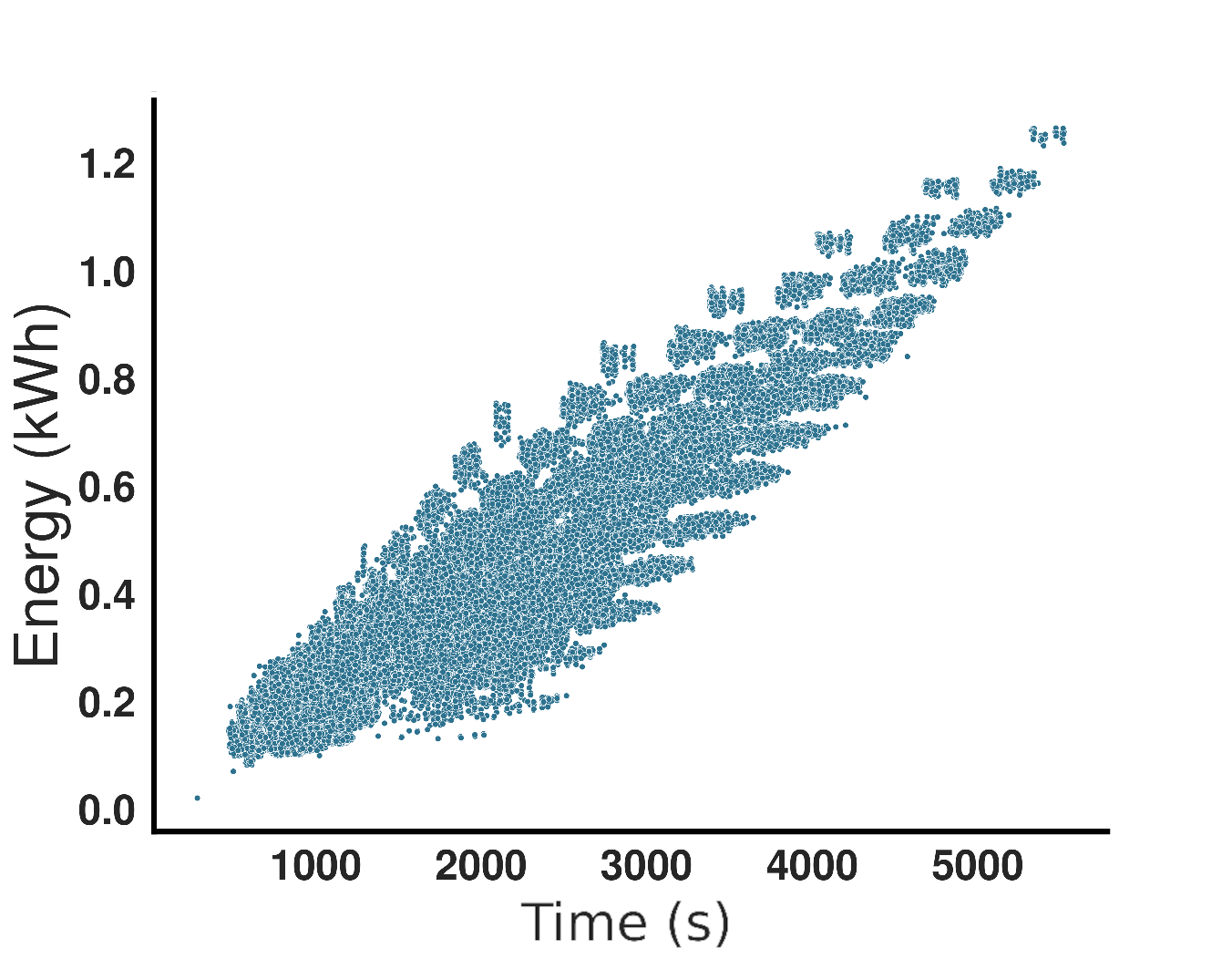}
        \caption{Energy consumption vs. training time
        }
        \label{fig:energy_vs_time}
    \end{subfigure}\hfill
    \begin{subfigure}[b]{0.46\columnwidth}
  \centering
    \includegraphics[width=0.8\textwidth]{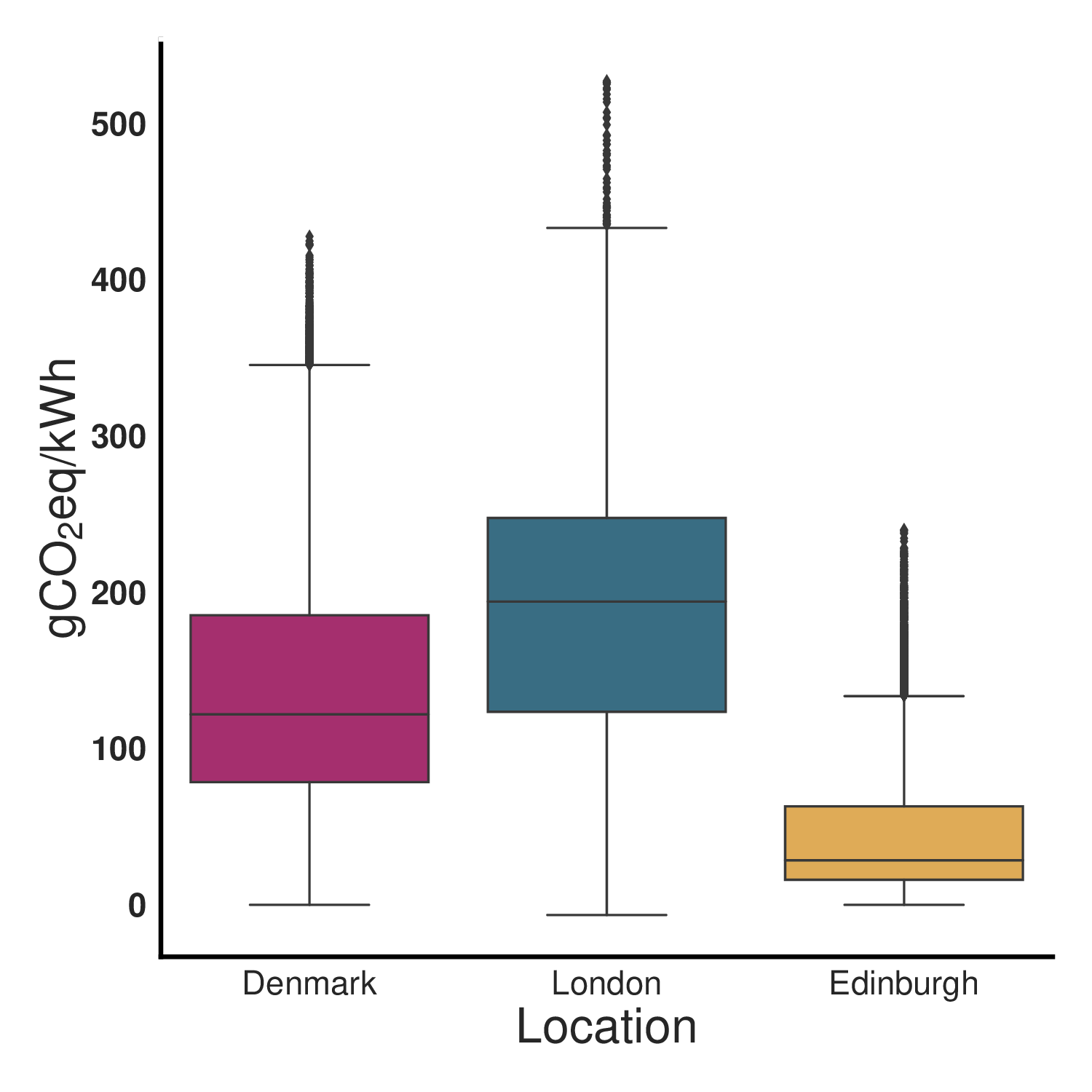}
    \caption{ Hourly carbon intensity
    }
    \label{fig:intensity_box_plot}
\end{subfigure}
    \caption{(a-c) Three plots demonstrating the discrepancy between different metrics of compute and energy, highlighting that changing one may not change another in kind. Note that each dot marker is a CNN model from the EC-NAS dataset~\cite{bakhtiarifard2022energy}. (d) Hourly carbon intensity in terms of \gcoeq{}/kWh over the time period 2019-2023 for three different regions: Denmark, London, and Edinburgh. The boxes show the median and interquartile range of carbon intensities; points outside the whiskers indicate outliers. Each region has vastly different distributions of carbon intensity, and all three are characterized by high variance with several peaks.}
\Description[Three plots demonstrating the discrepancy between different metrics of compute and energy as well as a plot showing hourly carbon intensity in three European regions]{Three plots demonstrating the discrepancy between different metrics of compute and energy. The first plot shows a large variation in terms of training time for equivalently sized models, demonstrating that model training time is not a strictly monotonically increasing function of model size. The second plot shows the same pattern 
for energy consumption vs.~the number of parameters and the third plot for energy consumption  vs.~training time. Furthermore, a forth plot visualizes the hourly carbon intensity in Denmark, London, and Edinburgh, showing a large variability.}
\end{figure}

\begin{itemize}
    \item Discrepancy 1: Compute efficiency $\neq$ energy efficiency $\neq$ carbon efficiency.
    \item Discrepancy 2: Efficiency has unexpected effects on operational emissions across the ML model life cycle.
    \item Discrepancy 3: Efficiency does not account for, and can potentially exacerbate, broader environmental impacts from hardware platforms.
\end{itemize}

Based on these we argue that to make ML more environmentally sustainable, it will be necessary to address the complexity resulting from the interaction of many factors which affect the sustainability of ML \textit{as a technology}. Here, ML ``as a technology'' considers not just the instruments of ML but also the social relations it induces, in the sense that ``technology is and does what people say it does and is''~\cite{heikkurinen2021sustainability}. In other words, ML as a technology includes ML systems and the people who use them: computation, ML model life cycles, human behavior, the supply chain, economic forces, and more. We posit that systems thinking, which provides a lens and framework with which to deal with complexity, offers a potential path towards accomplishing the goal of making ML as a technology environmentally sustainable~\cite{richmond1994system}. Systems thinking seeks to understand the relationship between the structure and behavior of complex systems, which can reveal unexpected effects arising from the interaction of the components which comprise the system. The discrepancies we describe in this paper about improving efficiency of ML and environmental sustainability are examples of such unexpected effects which could potentially be better characterized and mitigated through systems thinking. 
\begin{figure*}[t]
  \centering
    \includegraphics[width=0.85\linewidth]{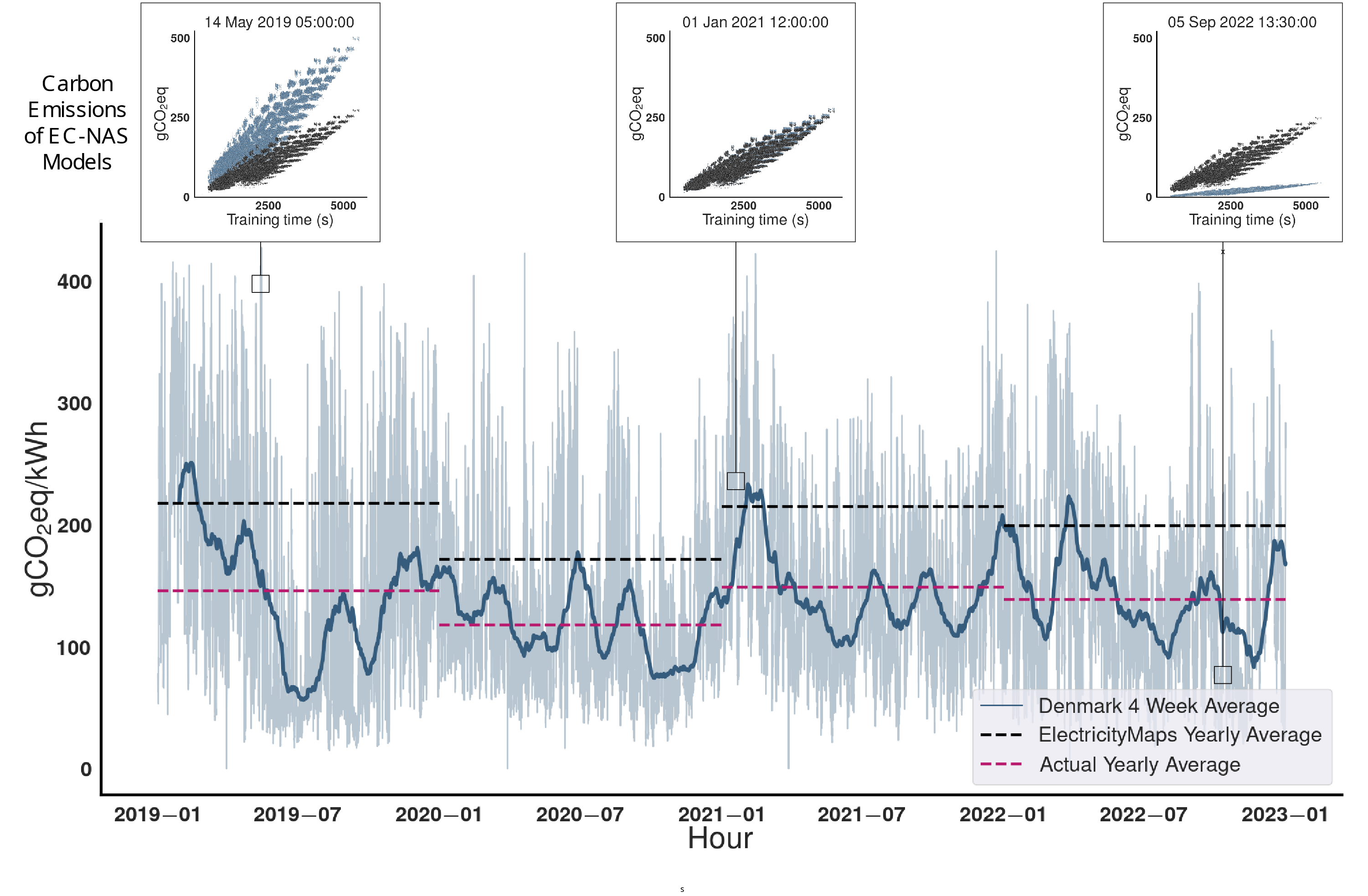}

    \caption{Monthly rolling average carbon intensity (thick blue line) and hourly carbon intensity (thin blue line) for Denmark between 2019 and 2023. To compare with estimates, we show the actual yearly average and yearly average carbon intensities for the same time period in Denmark. Subplots demonstrate how the selection of start time and which carbon intensity measure to use can result in vastly different observed operational emissions for the 423K models in the EC-NAS benchmark (blue points are using real-time carbon intensity, black points are using ElectricityMaps average intensity). Emissions are calculated by averaging the total energy consumption of each model over the selected time period, multiplying the energy consumption by the instantaneous carbon intensity in 5 minute intervals.
    }
    \label{fig:emissions}
\end{figure*}

\section{Discrepancy 1: Compute Efficiency $\neq$ Energy Efficiency $\neq$ Carbon Efficiency}
\label{sec:efficiency}
At face value it would appear that reducing compute would reduce energy consumption, which would in turn reduce carbon emissions. However, operational carbon emissions are a function of both energy and carbon intensity, which is dependent on time and location, and energy is a complex function of several factors which metrics of compute (e.g. FLOPS, number of parameters, and runtime) do not fully capture. As such, savings made in the amount of compute used in a model based on these metrics do not always translate to savings in energy due to e.g. the specifics of model architecture and hardware~\cite{DBLP:journals/corr/abs-2002-05651,DBLP:conf/nips/JeonK18,peng2023efficiency}. Furthermore, savings in energy consumption may not translate into savings in operational carbon emissions if one does not run their compute in locations and times where carbon intensity is low~\cite{DBLP:journals/corr/abs-2002-05651,anthony2020carbontracker,DBLP:conf/fat/DodgePCOSSLSDB22}\footnote{At any given time point, the energy mix used for electricity generation in the power grid can vary depending on several factors (availability of renewable sources, demand on the power grid, etc.). These factors influence the instantaneous carbon emissions of electricity production which is captured as {\em carbon intensity}.}. This discrepancy has been well documented in the literature, with multiple studies demonstrating and calling for a more holistic perspective on model efficiency~\cite{DBLP:journals/corr/abs-2002-05651,DBLP:conf/nips/JeonK18,DBLP:conf/ijcnn/QinZLZP18,DBLP:conf/hotcloud/YeungBFHG20,yarally2023uncovering,thompson2020computational,peng2023efficiency,kaack2022aligning}.

We present further evidence of the unintuitive effects of compute efficiency on operational emissions and energy.
We look at 423,624 models from the energy consumption aware neural architecture search (EC-NAS) benchmark dataset~\cite{bakhtiarifard2022energy} which contains training costs and performance metrics for all the models in a large space of convolutional neural networks (CNNs), including their training energy consumption.
We look directly at the commonly used measures of computational efficiency, namely model size (in number of trainable parameters) and training time in \autoref{fig:time_vs_params}, regional variations in carbon intensity (Denmark, Edinburgh, and London)\footnote{We query two publicly available sources which provide historical carbon emissions for Denmark in 5 minute intervals (\url{https://www.energidataservice.dk/}) and local regions in the UK for 30 minute intervals  (\url{https://carbonintensity.org.uk/})} in \autoref{fig:intensity_box_plot}, and the potential operational emissions of the EC-NAS models using {\em real-time} carbon intensity in \autoref{fig:emissions}.

Starting with \autoref{fig:time_vs_params}, similar to previous work~\cite{DBLP:journals/corr/abs-2002-05651,DBLP:journals/corr/abs-2210-06640} we see a large variation in terms of training time for equivalently sized models i.e. model training time is not a strictly monotonically increasing function of model size. This is further reflected in the energy consumption of each model versus the number of parameters, shown in \autoref{fig:energy_vs_params}, as well as the energy consumption of each model versus its training time, shown in \autoref{fig:energy_vs_time}. Hence, even within the same model type i.e. CNNs, the amount of compute compared to the amount of energy consumption is not always one-to-one. In the case of CNNs, for example, different operations and architecture choices which are not dependent on the number of parameters (e.g. batch/layer normalization, the use of residual and skip connections, the choice of activation function, etc.) lead to this discrepancy.

\begin{figure}[t]
  \centering
    \includegraphics[width=0.98\linewidth]{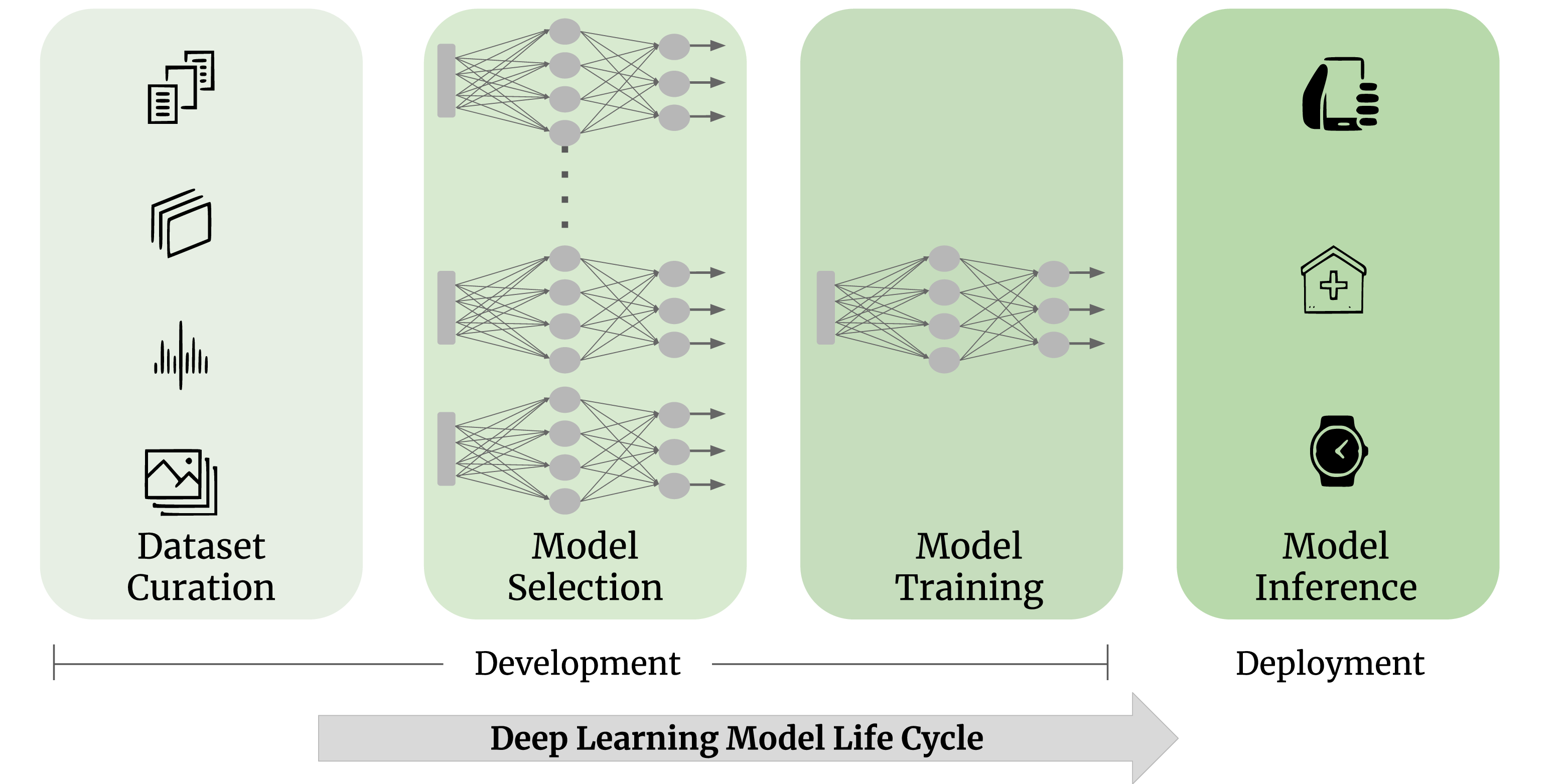}
    \caption{The Deep Learning model life cycle. The model development stage consists of data curation, model selection, and model training, while the deployment stage consists of the use of a model for inference in downstream applications and potentially retraining a model on new data.
    }
\Description[Illustration of the Deep Learning model life cycle]{The figure illustrates the Deep Learning model life cycle. It shows the development phase, starting with data set curation, followed by models selection and then model training, and the deployment phase in which the final model is used for inference.}
    \label{fig:lifecycle}
\end{figure}

Looking at carbon intensity (\autoref{fig:intensity_box_plot}), we see that each location has a vastly different average intensity, a large amount of variation, and several peaks as indicated by the number of outliers. Carbon intensity can change sporadically as a result of changing demand. ML jobs which could otherwise be run when carbon intensity is low have the potential to emit far more carbon than is necessary~\cite{DBLP:conf/fat/DodgePCOSSLSDB22}. This is reflected in \autoref{fig:emissions}, which highlights how real-time carbon intensity varies drastically both for the time of year and the time of day in Denmark, leading to vastly different expected operational emissions depending on when EC-NAS model training would be run. Further, we plot the average carbon intensity for Denmark retrieved from ElectricityMaps,\footnote{\url{https://app.electricitymaps.com/map}} an aggregator of real-time carbon intensity from around the world, as well as the actual average carbon intensity for each year, and compare this to the real-time carbon intensity, which turn out to be starkly different. As such, it is important to note that estimations of operational emissions, while useful and easier to compute than embodied emissions, can greatly over- or under-represent the true operational emissions.

The impact on operational emissions due to improvements in either compute or energy efficiency can often be different than expected. This is because variables such as runtime and number of parameters are not fully predictive of energy consumption, and energy consumption is not fully predictive of carbon emissions. 
Operational emissions at the level of compute are in fact a complex function of several variables, including e.g. the combination of model architecture and hardware platform~\cite{DBLP:journals/corr/abs-2002-05651,DBLP:conf/nips/JeonK18,peng2023efficiency} and when and where a model is run~\cite{DBLP:conf/fat/DodgePCOSSLSDB22}. In this regard, more work is needed to understand how these variables impact energy and operational carbon emissions in order to more effectively understand how to leverage efficiency. This also reveals that one should take care to actually measure compute, energy, and operational emissions to observe the impact of actions intended to reduce those emissions, for example, using one of the many available carbon tracking tools~\cite{DBLP:conf/emnlp/BannourGNL21,anthony2020carbontracker,DBLP:journals/corr/abs-2002-05651,schmidt2021codecarbon,budennyy2023eco2ai}; for comprehensive surveys of these tools see~\cite{budennyy2023eco2ai,DBLP:conf/emnlp/BannourGNL21,jay2023experimental,bouza2023estimate}.

\section{Discrepancy 2: Efficiency Across The Model Life Cycle}
Discrepancy 1 described the complexity arising from factors which influence operational emissions at the level of compute. Developing, producing, and using an ML system in practice results in many actions which require compute and energy and emit carbon.
Efficiency will impact the decisions one makes throughout the model life cycle, which will not always lead to reductions in carbon emissions. 
Here, we describe the unintuitive effects of efficiency on operational emissions when observed at the level of the model life cycle.%

\begin{figure*}[t]
  \centering
    \includegraphics[width=0.75\linewidth]{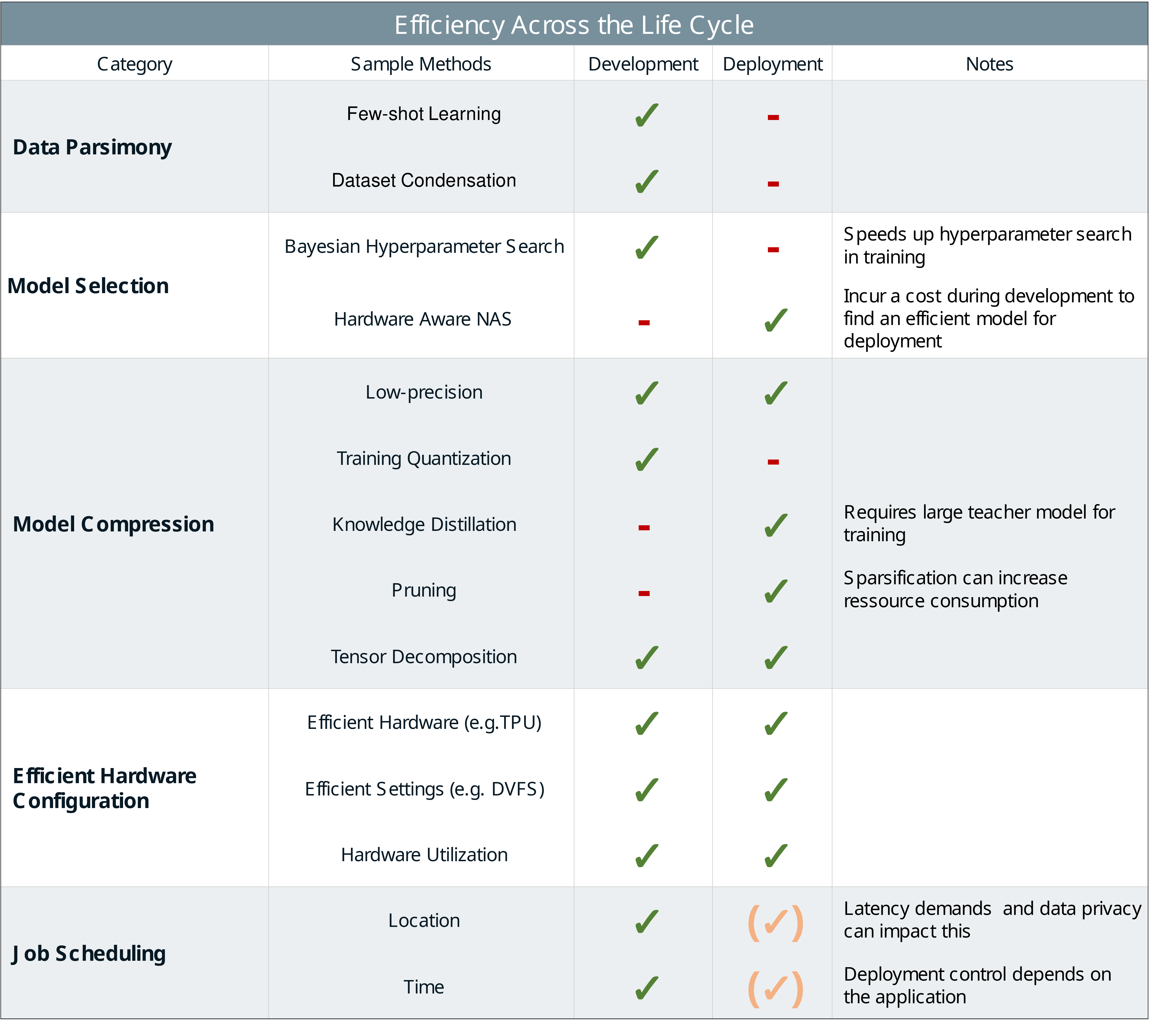}
    \caption{A sample of different ways to improve efficiency and whether or not it is targeted to development or deployment in a typical scenario (check means yes, dash means no, check within parentheses means it depends on the situation). Examples include data parsimony~\cite{DBLP:journals/csur/WangYKN20,DBLP:journals/corr/abs-2301-07014}, model selection~\cite{DBLP:journals/csur/RenXCHLCW21,DBLP:conf/ijcai/BenmezianeMONWW21}, model compression~\cite{DBLP:journals/pieee/DengLHSX20,DBLP:conf/icml/GuptaAGN15,DBLP:journals/corr/abs-2103-13630,DBLP:journals/ijcv/GouYMT21,DBLP:journals/ijon/LiangGWSZ21,wangexploring,DBLP:journals/corr/abs-2210-06640}, and hardware configuration~\cite{DBLP:conf/isca/JouppiYPPABBBBB17,DBLP:conf/cvpr/YangCS17,DBLP:conf/sc/AllenG16,yarally2023uncovering,DBLP:conf/bdcloud/LiCBZ16} for energy efficiency, and job scheduling~\cite{DBLP:conf/fat/DodgePCOSSLSDB22} for carbon efficiency.
    }

    \label{fig:efficiency_methods}
\end{figure*}

The model life cycle is generally broken down into two primary stages: development and deployment (see \autoref{fig:lifecycle}). The split in compute, energy, and operational emissions between development and deployment depends on several factors: for example how large a given model is, what algorithms one uses to design and find a suitable model, how readily a developed model is adopted by end users, and how long that model is used for.
In practice, deployment can end up constituting 90\% of the compute of a model over its lifetime~\cite{DBLP:conf/mlsys/WuRGAAMCBHBGGOM22,DBLP:journals/corr/abs-2104-10350,DBLP:journals/computer/PattersonGHLLMR22,openai_2018,leopold_2019}, which can lead to much greater operational emissions during deployment. This is critical, as many methods that are advertised as ``efficient'' are mainly applicable to only one part of the model life cycle as opposed to both, and may in fact incur a net increased cost in the end. Several examples of these actions are provided in \autoref{fig:efficiency_methods}, labeled by whether they are intended to reduce operational emissions in the development or deployment stages. Applying or abstaining from the use of efficient methods can thus potentially have a far-reaching impact on the total operational emissions of a model over its life cycle. For example, job scheduling allows one to reduce operational emissions during training by selecting to train one's models at times and locations with lower carbon intensity~\cite{DBLP:conf/fat/DodgePCOSSLSDB22}. 
AI systems may offer the opportunity to decouple where a service is used and where most energy is consumed.
However, job scheduling is not always a viable option, as the ability to select where and when to run may be limited due to constraints on how the trained models are used (e.g. when deployment latency and on-demand use or privacy are of concern). As another example, large development emissions can be incurred in order to save during deployment, such as with hardware-aware NAS~\cite{DBLP:conf/ijcai/BenmezianeMONWW21} and large, sparsely activated models~\cite{DBLP:conf/iclr/CaiGWZH20,DBLP:conf/icml/DuHDTLXKZYFZFBZ22}. How to holistically minimize operational emissions over the entire model life cycle as such is an open question, and addressing it requires being able to characterize the operational emissions resulting from multiple decisions over time.

Furthermore, attempts to reduce operational emissions via efficiency may not succeed in practice, as theoretical reductions in operational emissions (e.g., through the deployment of efficient models) can eventually result in greater emissions in practice. 
It is well documented that energy and carbon mitigation strategies are subject to rebound effects~\cite{dutschke2021rebound,font2022rebound,widdicks2023systems} (a.k.a. Jevons paradox~\cite{alcott2005jevons}) which occur when the observed reduction in carbon emissions due to an improvement in efficiency is not as significant as the expected reduction, or could actually result in an increase in emissions. It has indeed been noted in the literature on the environmental sustainability of ML that new ML models introduced in industry which improve the efficiency of those industries can potentially lead to an increase in carbon emissions. Examples include ML systems which increase the production of goods at a manufacturing plant and ML powered autonomous vehicles leading to more individual travel~\cite{kaack2022aligning}. This also extends to ML itself, as making models more efficient can rebound via increased usage of those models. The rebound effect has been documented at multiple large companies with respect to energy consumption from ML systems~\cite{DBLP:conf/mlsys/WuRGAAMCBHBGGOM22,DBLP:journals/computer/PattersonGHLLMR22}. It occurs for a number of reasons, but is largely facilitated by economic, psychological, and behavioral factors which accompany efficiency improvements. 

It is easy to identify plausible examples of rebound effects in ML which many practitioners may find relatable: for example a practitioner makes an improvement in the compute efficiency of a model they are developing, which allows them to train that model on a single GPU device as opposed to two, and in half the time. This gain in efficiency offers the possibility of a larger scale of experimentation.%
The practitioner may now take the opportunity to train longer and on more data and   
decide to explore a broader range of hyperparameters, to determine the settings they will use to train their final model. This ultimately takes longer, consumes more energy, and produces more operational emissions than if they had performed a more limited random hyperparameter search with their original, less efficient model. This type of behavior can be attributed to perceived ``attenuated consequences'' from making the model more efficient~\cite{santarius2018technological}.

As such, operational emissions throughout the model life cycle can be particularly difficult to predict, as they are largely driven by behavior stemming from both a lack of awareness and competing incentives~\cite{widdicks2023systems}. More concretely, this can be a lack of awareness of what aspects of the model life cycle a particular efficiency improvement is targeting, behavior which leads to significantly more compute over time
~\cite{dutschke2018moral,santarius2018technological}, incentives to scale up in order to improve accuracy and serve a larger user base, and more. The net effect is that improved efficiency does not mean that operational emissions across the life cycle will reduce, in some cases it can lead to further increases. Thus, in addition to the factors we discussed previously at the level of compute, we must reckon with different factors at the level of model life cycles which affect operational emissions in order to move towards the goal of reducing them. This calls for both technical and non-technical (e.g., regulatory) solutions.

\section{Discrepancy 3: Efficiency and Platforms}

As reviewed with both compute and the model life cycle, efficiency alone does not fully address operational carbon emissions i.e. those due to compute. Computing platforms (the hardware and infrastructure on which ML compute runs), come with their own set of environmental impacts including but not limited to carbon emissions. These impacts are diverse and highly distributed among many processes and people, making them complex in and of themselves, and have the potential to worsen going forward as ML becomes more widely adopted~\cite{robbins2022our}. Efficiency can have both positive and negative impacts on this; on the one hand reducing the compute and energy needs of hardware and on the other hand facilitating the greater use and manufacture of existing and emerging hardware platforms~\cite{kaack2022aligning,robbins2022our,batra2018artificial}.  In light of this, it is becoming increasingly important to account for the environmental impacts of ML platforms and the factors which give rise to them.

Manufacturing the devices on which ML systems operate requires the mining of different materials (e.g. critical minerals), yielding multiple pollutants and hazardous products such as radioactive and toxic chemical components~\cite{balaram2019rare,IEA2023}. Poor mining practices can lead such chemicals to enter food and water supplies and cause downstream health impacts~\cite{nayar_2021}. The mining of resources such as gold, nickel, copper, and other critical minerals additionally contribute significantly to deforestation~\cite{wwf2022}, threaten to worsen the effects of climate change, impact biodiversity and critical ecosystems such as those in the Amazon~\cite{siqueira2020keep}, and harm Indigenous communities~\cite{IEA2023}. 
It is currently  unclear what the contribution of ML systems is to these impacts as data describing them is lacking, but they are known to be significant in the ICT sector as a whole~\cite{robbins2022our,edahbi2019environmental}. 

Additionally, the mining and device manufacturing process result in their own carbon emissions (a.k.a embodied emissions). These embodied emissions can vary greatly, where it has been estimated that they account for approximately 10\% of total emissions in data centers and 40-80\% of total emissions for devices at the edge such as mobile phones and sensors which collect data~\cite{malmodin2018energy,whitehead2015life,masanet2013characteristics}. A significant portion of a model's total carbon footprint can come from embodied emissions. For example, Luccioni et al.~\cite{DBLP:journals/corr/abs-2211-02001} estimate that the embodied emissions from training BLOOM~\cite{bloom}, a 176B parameter large language model, constituted 22\% of its total emissions (11.2 tons \co{}eq). Projecting forward, it has been estimated that embodied emissions may become the dominant source of emissions both within ML~\cite{DBLP:conf/mlsys/WuRGAAMCBHBGGOM22} and in the ICT sector as a whole~\cite{gupta2022chasing}, partially as a result of the rise of edge compute running ML systems.

Furthermore, much of ML compute, particularly with the emerging large deep learning models~\cite{bloom,DBLP:conf/nips/BrownMRSKDNSSAA20}, is performed in data centers. Data centers require a significant amount of water for electricity generation and cooling; ML systems are playing an increasingly large role in this water consumption ~\cite{mytton2021data,DBLP:journals/corr/abs-2304-03271}. For example, Li et al. (2023)~\cite{DBLP:journals/corr/abs-2304-03271} estimate that the water consumption from GPT-3~\cite{DBLP:conf/nips/BrownMRSKDNSSAA20}, another large language model with 175B parameters, required 700,000 liters of clean fresh water to train. Accounting for this is important as this increased water usage can contribute to water scarcity. This is becoming an increasingly salient issue with the effects of climate change, making droughts more common\footnote{\url{https://www.unwater.org/water-facts/water-scarcity}}, and in some cases large data centers can compete with local communities for clean freshwater resources~\cite{bast2022four}.

Finally, at their end of life, devices will be either recycled, repurposed, or disposed of, where repurposing and disposal result in e-waste~\cite{vidal2013toxic,wang2024waste}. 
Environmental impacts from this relate to the physical dumping of e-waste on land. With so much waste, hazardous chemicals can leak into the land and water supplies~\cite{nayar_2021} and affects local biodiversity. Furthermore, e-waste sites offer a source of livelihood for many communities who scavenge the digital components for minerals to sell. Minerals are hard to recover, and as such must be extracted by open-air burning of waste and the use of acid baths. Not only does this have catastrophic affects on these communities' health, but it can also lead to air pollution, and the further release of toxic chemicals into the land and water. Similar to the impacts of mining, the contribution of ML to the impacts from e-waste are not well understood.

Compute and energy efficiency can play a role in helping to limit ML's need for and use of hardware, but will not eliminate it nor its associated environmental impacts. At the level of data centers, efficiency has helped to limit energy consumption rising at the same pace as compute loads in recent years~\cite{masanet2020recalibrating}. Additionally, typical server refresh times, where devices reach end of life (e-waste) and new devices are purchased and installed (resulting in embodied emissions and all of the impacts from device manufacturing), appear to be slowing, potentially with the help of increased device energy efficiency~\cite{uptime2022pue}. However, device energy efficiency is also slowing, in line with the slowing of Moore's Law~\cite{shankar2022trends}, so it is not clear if this trend will continue. 
Additionally, the power density of data centers (i.e. the amount of power drawn per server rack as a result of packing more compute into less space) has also been increasing in recent years, which can lead to an increased need for liquid cooling to stave off heat (thus consuming more water)~\cite{uptime2022pue}. The usage of ML hardware accelerators such as GPUs may be contributing to this~\cite{kaack2022aligning}. Additionally, as with efficiency across the life cycle, efficiency at the level of hardware could potentially result in rebound effects as hardware becomes cheaper, leading to increased demand~\cite{gossart2015rebound}. Indeed there has been increasing demand for ML hardware in recent years~\cite{batra2018artificial} despite improvements in efficiency~\cite{hernandez2020measuring,desislavov2021compute,DBLP:journals/pieee/DengLHSX20}, which is likely to continue going forward. %
This is particularly the case for edge devices, as improvements in compute and energy efficiency enable more ML compute to be performed outside of large data centers. The use of these devices is desirable in order to reduce latency and operational energy demands and thus cost. As such, the use of these devices for ML applications is expected to grow rapidly in the coming years~\cite{robbins2022our,DBLP:conf/hpca/WuBCCCDHIJJLLLQ19}. This has the potential to facilitate rebound effects in their operational energy consumption and carbon emissions as a result of their increased efficiency~\cite{DBLP:conf/mlsys/WuRGAAMCBHBGGOM22}. Additionally, the broader environmental impacts of device manufacture will potentially worsen if not accounted for and mitigated.

Given this, efficiency at the level of platforms is limited by both the slowing of hardware energy efficiency~\cite{shankar2022trends} as well as behavioral limits with the rebound effect~\cite{gossart2015rebound}.  
Worse, even accounting for the environmental impacts of platforms as a result of ML is currently difficult due to the complexity of factors which contribute to them and/or a lack of transparency~\cite{malmodin2018energy,DBLP:journals/corr/abs-2304-03271}. As such, platforms add a significant amount of complexity to the problem of making ML environmentally sustainable. Addressing this, as well as the impacts from compute across the model life cycle, will benefit from understanding and managing this complexity. In this light, efficiency is only a partial solution.

\section{Beyond Efficiency: Systems Thinking}

While we are critical of efficiency throughout this perspective, we note that it is still important as it can \textit{help} eliminate the environmental impact of ML systems. Thus, we encourage the community to foster a more honest and realistic discourse around efficiency in ML by (1) being precise about what is efficient when describing ``efficiency'' and (2) being wary of conflating efficiency with environmental sustainability as a whole. The discrepancies described in this perspective are intended to elucidate \textit{why} efficiency is not enough to achieve the goal of making ML as a technology environmentally sustainable. %
We see efficiency as one aspect to improve the environmental sustainability of ML which interacts with several variables at multiple levels. Individual agency to enact change becomes more difficult due to increasing complexity, thus necessitating more collaboration and cooperation.

This complexity leads to other systemic issues beyond the unintuitive effects of efficiency. For example, depending on what factors are chosen to be measured and how values such as the efficiency of data centers, embodied emissions, and carbon intensity are determined, one can conclude either that the carbon footprint of ML training will plateau and shrink~\cite{DBLP:journals/computer/PattersonGHLLMR22} or that the observed exponential increase in the carbon footprint of ML training~\cite{DBLP:journals/corr/abs-2302-08476} will continue in the near future. 
These issues persist at the level of individual models, exemplified in the difference in reported carbon emissions of Evolved Transformer~\cite{DBLP:conf/icml/SoLL19} by Strubell et al.~\cite{DBLP:conf/acl/StrubellGM19} and Patterson et al.~\cite{DBLP:journals/computer/PattersonGHLLMR22}. 
The goal of the paper from Strubell et al.~was to characterize the carbon emissions of modern ML circa 2019; as one component of this, they were forced to estimate some quantities needed to calculate the emissions of the model selection stage for Evolved Transformer (due to lack of transparency and reporting of these emissions in the Evolved Transformer paper),
including variables related to the compute itself and variables related to the infrastructure used to run the compute. Three years later, Patterson et al. then 
argued that the previous estimate was approximately $88\times$ {too high}\footnote{3.2 tons \co{}e vs. 284 tons \co{}e}  when considering the actual settings used for model selection. These differences arise from a lack of transparency of critical data (e.g. embodied emissions) and misalignment between ideas of what factors in ML to consider when measuring environmental impacts. This, in addition to the discrepancies discussed previously, illuminates the need for a new way to approach the environmental sustainability of ML as a technology which is more holistic and effective.

One way is to adopt systems thinking~\cite{amissah2020systems,meadows2008thinking}. Systems thinking is a well established field of study~\cite{richmond1994system} which has been successfully applied in several areas including engineering, management, computer science, and sustainability~\cite{hofman2018could,garrity2018using,widdicks2023systems}. It seeks to understand the relationship between the structure and behavior of complex systems: ``interconnected sets of elements which are coherently organized in a way that achieves something''~\cite{meadows2008thinking}. These complex systems are found everywhere: the bodies of living things, cities, companies, computer systems, etc. A key feature of systems thinking is the insight that complex systems are more than the sum of their parts. This is revealed through the systems lens, which looks at the behavior of the entire system as a whole, relating the components of the system to each other through causal feedback loops. This can reveal previously unobserved and unexpected behavior, meaning that the ``something'' which a system achieves might not be that which was intended by its designers~\cite{amissah2020systems}. This contrasts with an approach that breaks a larger system down into more easily studied components, which obfuscates this behavior~\cite{DBLP:journals/corr/abs-2104-10350,DBLP:journals/computer/PattersonGHLLMR22}. Essentially, systems thinking is a conceptual shift from seeing how individual causes give rise to behavior (e.g. a person reduces their carbon footprint by taking the bus instead of driving a car) to seeing how systems themselves behave (e.g. carbon emissions are produced by the transportation system, in which people, buses, and cars are a part).

How can systems thinking bridge the gap between efficiency and the environmental sustainability of ML as a technology? Consider a standard practice in ML for improving model training and inference efficiency: using mixed precision, where the number of bits used in computations is dynamically adjusted~\cite{micikevicius2018mixed}. Use of mixed precision computations should reduce the energy consumption of an ML model and thus operational carbon emissions when observed in isolation.\footnote{Mixed precision is the practice of switching between different quantization levels for the ML model weights and other intermediate estimates. This has shown to improve the computational efficiency with little or no reduction in performance.} Just the use of mixed precision is a sufficient condition for achieving ``efficiency.'' However, systems thinking invites us to observe and understand the behavior which arises through the systems lens, and an action such as using mixed precision interacts with many variables affecting ML environmental sustainability, thus producing potentially unintuitive effects on variables such as carbon emissions. 
One can consider how reducing the bit precision of a model interacts with, for example, its speed, which can in turn influence how much experimentation one chooses to perform in order to find the best model, facilitating the rebound effect (discrepancy 2). Going further, one can account for changes in the model's accuracy, which, combined with speed, can influence how frequently that system can be expected to be used, thus affecting operational emissions over time. One can then determine how each of these factors will influence the amount of hardware infrastructure required to support the downstream use of that model, as well as the type of hardware (e.g. edge devices vs. cloud data centers) likely to be used as a result of improved algorithmic efficiency (yielding discrepancy 3). Thus, systems thinking is intended to reveal how a seemingly isolated change such as using mixed precision inevitably ``releases or suppresses a behavior that is latent within the structure'' of the system itself~\cite{meadows2008thinking}, where the ``system'' in this case encapsulates ML compute, life cycles, and platforms.

Importantly, understanding such systems and their tendency towards particular behaviors can enable us to identify ways to both make the best use of the tools we have (e.g. efficiency) and discover other effective leverage points (e.g. socio-economic regulation) to enact a desired change (e.g. reduce carbon emissions). This is becoming more critical with ML as a technology in order to prevent undesirable systemic effects such as the ``lock-in'' of environmentally damaging behaviors~\cite{robbins2022our}. In such a scenario, ``prior decisions constrain future paths'' towards reducing environmental impacts due to the economic, social, and political conditions which cause a system to maintain a particular set of behaviors. This could occur in the case where groups of people or industries become dependent on the use of ML systems, but the socio-political regulations and technological developments are not in place to ensure that the use of these systems does not cause irrevocable damage to the environment. Greater measures than efficiency are needed in order to prevent this, and the time to start working on them is now.

Furthermore, systems thinking aims to understand the interconnections in a system ``in such a way as to achieve a desired purpose''~\cite{amissah2020systems}. Thus, systems thinking has the potential to help move towards a ``desired purpose'' such as aligning ML as a technology with the SDGs~\cite{DBLP:journals/aiethics/Wynsberghe21,kaack2022aligning}. This enables us to consider not just the environmental sustainability \textit{of} ML, but also ML \textit{for} environmental sustainability~\cite{DBLP:journals/csur/RolnickDKKLSRMJ23}, the relationship of ML as a technology with economic and social sustainability, and how these areas are connected. With respect to ML for environmental sustainability, ML can help optimize processes in many areas, as well as advance environmental sciences~(e.g., \cite{tuia2022perspectives,mugabowindekwe2023nation}), leading to a net positive environmental impact. As a concrete example, it has been estimated in the construction sector that ``widespread deployment of active controls, assuming limited rebound effects, would save up to 65 PWh cumulatively to 2040, or twice the energy consumed by the entire buildings sector in 2017''~\cite{iea2017}. These active controls come in the form of e.g. smart thermostats and lighting which can ensure effective use of energy, both of which are improved with the use of ML. When it comes to economic and social sustainability, the increased adoption of ML and choices about how to implement and deploy ML systems can have impacts on these areas. For example, the environmentally sustainable choice to use low carbon data centers~\cite{DBLP:journals/corr/abs-2104-10350} requires thinking about social sustainability due to the potential for data privacy and surveillance issues~\cite{lyon2014surveillance}. Considerations such as these should be balanced against those which seek to make ML as a technology more environmentally sustainable.

Given the complexity and cross-disciplinary nature of reaching a systems level understanding of ML as a technology and its impacts in practice, interdisciplinary collaboration is key. This has been done with initial work on identifying factors which affect ML sustainability holistically~\cite{kaack2022aligning,ligozat2022unraveling}, developing governance frameworks~\cite{rohde2023broadening}, developing reporting frameworks~\cite{DBLP:journals/corr/abs-2002-05651,DBLP:conf/emnlp/HershcovichWKBL22,DBLP:journals/corr/abs-1910-09700}, revealing the hidden costs of ML use~\cite{DBLP:conf/fat/DodgePCOSSLSDB22,DBLP:journals/corr/abs-2304-03271}, and more. A necessary step will be to foster more dialogue around these impacts: what impacts to measure, how to measure them, and what influences them. This can help us to model ``the rules of the game'' i.e. how these impacts arise as a result of system level behavior. 
Important questions then arise: what are effective interventions for changing the way ML as a technology operates for better? Who can and should be involved in implementing these interventions? What \textit{negative} impacts do we want to limit and what \textit{positive} impacts do we want to encourage from ML? A systems level understanding of ML as a technology offers a more informed way to explore these questions.

\section{Conclusion}

With respect to environmental sustainability, the ML community currently relies heavily on efficiency as the solution of choice~\cite{DBLP:journals/corr/abs-2301-11047,DBLP:conf/mlsys/WuRGAAMCBHBGGOM22,DBLP:journals/computer/PattersonGHLLMR22,DBLP:journals/corr/abs-2210-06640}. This is not without sensible motivation: efficiency can reduce carbon emissions, it is often easy to measure and implement, it lends itself as a metric by which one can compare different systems and methods, it can be deployed in many ways without requiring coordination between large groups of people, and it can help to serve other goals such as making ML systems faster and cheaper to operate. However, as ML systems are becoming increasingly prevalent~\cite{robbins2022our}, it is incumbent on us to move beyond the dominating focus on efficiency and to cultivate a more nuanced view of the environmental impact of ML as a technology and ways to reduce it. In this paper we demonstrate why this is the case by describing three discrepancies between efficiency and the goal of environmentally sustainable ML, and propose systems thinking as a way to move beyond efficiency. The discrepancies include: compute, energy, and carbon are not equivalent, operational emissions across the ML model life cycle are affected by efficiency in unexpected ways, and efficiency alone is not enough to address the broader environmental impact of platforms.
We thus illuminate opportunities for new research, policy, and practice which can improve the environmental sustainability of ML as a technology {\em holistically}.

\begin{acks}
We acknowledge Pedram Bakhtiarifard for access to and help with the EC-NAS benchmark data. DW, CI and RS are partly funded by the European Union's Horizon Europe research and innovation programme under grant agreements No.~101070284 and No.~101070408. 
CI acknowledges by the Pioneer Centre for AI, DNRF grant number P1.
GS would like to acknowledge Wellcome Foundation (grant number 222180/Z/20/Z).
\end{acks}

\printbibliography

\end{document}